\documentclass[10pt,twocolumn,letterpaper]{article}

\usepackage{cvpr}              
\definecolor{cvprblue}{rgb}{0.21,0.49,0.74}
\usepackage[pagebackref,breaklinks,colorlinks,allcolors=cvprblue]{hyperref}
\usepackage{tabularx} 
\usepackage[utf8]{inputenc} 
\usepackage{booktabs}
\usepackage{xcolor}
\usepackage{pifont}    
\usepackage{caption}
\usepackage{geometry}
\usepackage{graphicx}
\usepackage{multirow}
\usepackage[table]{xcolor}
\usepackage{float}  
\usepackage[cache=true, cachedir=.]{minted}
\usepackage{float}
\usepackage{placeins}
\usepackage[T1]{fontenc}    
\usepackage[utf8]{inputenc} 
\usepackage[accsupp]{axessibility}  

\newcommand{\cmark}{\textcolor{green!50!black}{\ding{51}}} 
\newcommand{\xmark}{\textcolor{red}{\ding{55}}}           
\newcommand{\samethanks}[1]{\footnotemark[#1]}

\def\paperID{39596} 
\def\confName{CVPR}
\def\confYear{2026}

\title{Omni IIE Bench: Benchmarking the Practical Capabilities of Image Editing Models}

\author{
Yujia Yang$^{1}$\thanks{Equal contribution.},
Yuanxiang Wang$^1$, Zhenyu Guan$^1$, Tiankun Yang$^1$, Chenxi Bao$^1$\\
Haopeng Jin$^2$, Jinwen Luo$^2$, Xinyu Zuo$^2$, Lisheng Duan$^2$\\
Haijin Liang$^2$, Jin Ma$^2$, Xinming Wang$^2$, Ruiwen Tao$^2$,
Hongzhu Yi$^{1}$\samethanks{1} \thanks{Corresponding Author.}\\
$^1$University of Chinese Academy of Sciences, 
$^2$Tencent\\
\url{https://github.com/Young-2000/OmniIIEBench}
}

\begin{document}
\maketitle
\begin{abstract}
While Instruction-based Image Editing (IIE) has achieved significant progress, existing benchmarks pursue task breadth via mixed evaluations. This paradigm obscures a critical failure mode crucial in professional applications: the inconsistent performance of models across tasks of varying semantic scales. To address this gap, we introduce \textbf{Omni IIE Bench}, a high-quality, human-annotated benchmark specifically designed to diagnose the editing consistency of IIE models in practical application scenarios. Omni IIE Bench features an innovative dual-track diagnostic design: (1) \textbf{Single-turn Consistency}, comprising shared-context task pairs of attribute modification and entity replacement; and (2) \textbf{Multi-turn Coordination}, involving continuous dialogue tasks that traverse semantic scales. The benchmark is constructed via an \textbf{exceptionally rigorous} multi-stage human filtering process, incorporating a quality standard enforced by computer vision graduate students and an industry relevance review conducted by professional designers. We perform a comprehensive evaluation of 8 mainstream IIE models using Omni IIE Bench. Our analysis quantifies, for the first time, a prevalent performance gap: nearly all models exhibit a significant performance degradation when transitioning from low-semantic-scale to high-semantic-scale tasks. Omni IIE Bench provides critical diagnostic tools and insights for the development of next-generation, more reliable, and stable IIE models.
\end{abstract}
\section{Introduction}
\label{sec:intro}
\begin{figure*}[h]
    \centering
    \includegraphics[width=0.9\linewidth]{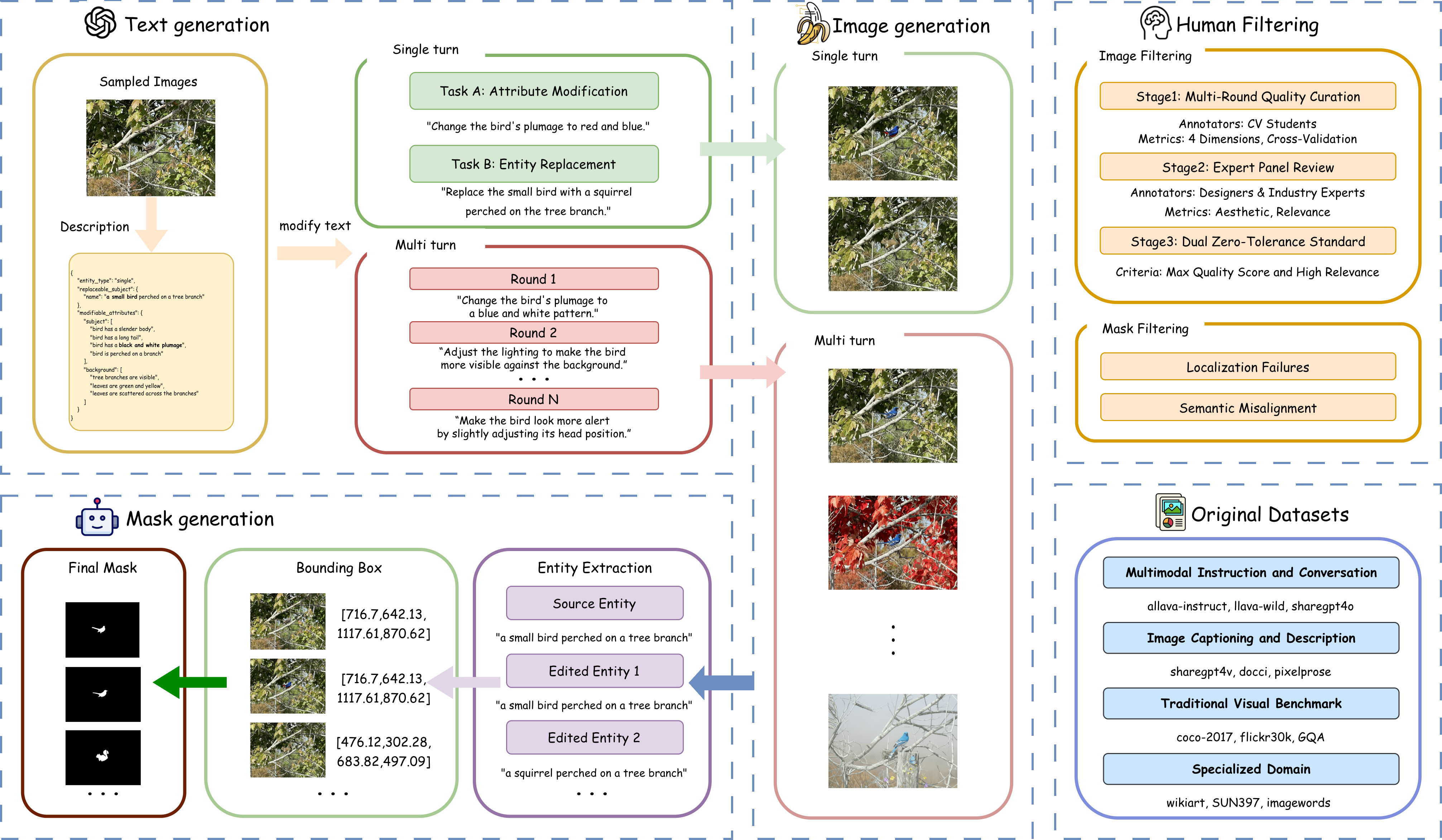}
    \caption{Omni IIE Bench collects seed images from 12 datasets and uses \textit{GPT-4o} to generate image descriptions. The descriptions are then randomly modified by \textit{GPT-4o}. For single-turn dialogue generation, semantic scales are done at low and high levels; for multi-turn dialogue generation, modifications are randomly interleaved between the two levels. After that, the original descriptions, modified descriptions, and original images are input into \textit{Nano Banana} for image generation, and the results are processed with \textit{GroundingDINO} and \textit{SAM} to obtain masks. Finally, all generated images and masks undergo strict manual review.}
    \label{fig:overview}
\end{figure*}
In recent years, the capabilities of multimodal image editing models \cite{batifol2025flux, brooks2023instructpix2pix, fu2024mgie, hui2024hq, wu2025qwenimagetechnicalreport, zhang2024hive, zhang2025context, liu2025step1x-edit, wang2025multimodal} have been continuously improving, driving the development of Instruction-based Image Editing (IIE) tasks. IIE models enable users to iteratively manipulate image content through natural language. This simple and intuitive editing paradigm greatly enhances the flexibility and interactivity of visual creation, rapidly becoming a core tool in the design domain.

Early IIE benchmarks primarily focused on single-turn dialogues \cite{ma2024i2ebench, wang2023imagen, basu2023editval, sheynin2024emu, Yu_2025_CVPR, wang2024mtubenchmultigranularitytoolusebenchmark, yu2025browseragent, yi2026rpo}, evaluating whether models could produce high-quality edits based on a single instruction. During this stage, editing instructions were mainly categorized into attribute modifications and entity modifications. However, single-turn settings overlook the importance of multi-turn interactions in real design workflows.

Recent studies have begun to explore multi-turn editing \cite{jia2025compbench, ye2025imgedit, zhou2025multi, zhang2023magicbrush, wu2025collabllmpassiverespondersactive, yu2026shotfinder, xie2025more}, but they are typically limited to 2–3 rounds of dialogue. In contrast, in practical design processes, designers tend to perform edits frequently and progressively, refining images through multiple iterations. Therefore, 2–3 rounds of interaction are insufficient to accurately assess a model’s true practical capability.

More importantly, existing IIE benchmarks have not been validated by experienced designers, leading to a noticeable gap between idealized benchmark performance and practical applicability.

To systematically evaluate the large-scale applicability of different IIE models in practical scenarios, it is thus essential to establish a diagnostic benchmark grounded in real design practice. To address these challenges, this paper proposes \textbf{Omni IIE Bench} — a rigorously human-evaluated benchmark designed to diagnose the practical performance of IIE models. The main contributions of this work are as follows:\\
\textbf{1. High-quality evaluation dataset:}
We propose a high-quality evaluation dataset consisting of both single-turn and multi-turn dialogue data. The single-turn dataset contains 1,725 image editing dialogues, while the multi-turn dataset includes 260 multi-round dialogues, with up to 16 turns per dialogue. The dataset was constructed with extensive human involvement and a rigorous data filtering process to ensure quality and reliability.\\
\textbf{2. Decoupled and robust evaluation methodology:}
We introduce a decoupled diagnostic evaluation framework that assesses model performance along three key dimensions: global image quality, decoupled regional fidelity, and instruction compliance. This framework provides a more interpretable and fine-grained evaluation of editing performance than traditional single-score methods.\\
\textbf{3. Systematic diagnostic analysis of mainstream IIE models:}
Using our proposed benchmark, we conduct a comprehensive evaluation of several mainstream IIE models, revealing their actual performance in practical scenarios. Furthermore, we perform human validation of the evaluation results, which show a high degree of consistency with automated assessments, demonstrating the effectiveness and reliability of Omni IIE Bench.

\section{Related Work}
\label{sec:related work}
\paragraph{Single-turn Evaluation Benchmarks:} Existing single-turn benchmarks tend to pursue task coverage by evaluating a mixture of tasks, thereby testing overall model capability \cite{ma2024i2ebench, wang2023imagen, basu2023editval, sheynin2024emu, zhang2025dynamic, wang2025hitchhiker, xie2025zeroes}. However, this mixed evaluation paradigm obscures the specific performance differences of models across different semantic-scale tasks. I2EBench \cite{ma2024i2ebench}, for example, attempts to explicitly separate \emph{high-level} and \emph{low-level} edits through a hierarchical design, which underscores the importance of the semantic scale concept. Yet, it also adopts an isolated checklist-style evaluation, assessing and reporting performance separately for different task levels, without diagnosing model stability when handling cross-scale task switching in the same image context.

\paragraph{Multi-turn Evaluation Benchmarks:} Other multi-turn benchmarks focus on evaluating \emph{temporal consistency} during the editing process \cite{jia2025compbench, ye2025imgedit, zhou2025multi, zhang2023magicbrush,zong2025jtcse, wang2025mr, zong2025tncse}. Although CompBench \cite{jia2025compbench} supports multi-turn editing and distinguishes four dimensions in its instruction decoupling strategy, it still reports each task score as an isolated metric, failing to quantify model consistency between low and high semantic scale tasks. ImgEdit-Bench \cite{ye2025imgedit} emphasizes assessing content memory, content understanding, and version rollback abilities. Similarly, MUCIE \cite{zhou2025multi} concentrates on instruction retention and reasoning memory to mitigate error accumulation. While these studies are crucial for interactive editing, they do not consider model stability and coordination when the semantic scale of editing tasks dynamically changes.

\begin{table*}[ht]
\centering
\caption{Comparison of IIE Benchmarks on Key Diagnostic Dimensions.}
\small
\renewcommand{\arraystretch}{0.7} 
\label{tab:compare}
\resizebox{\textwidth}{!}{%
\begin{tabular}{l|c|c|c|c|c}

\toprule
\textbf{Benchmark} & \textbf{Human Verified} & \textbf{Provides Mask} & \textbf{Practical Scenarios} & \textbf{Semantic Scale} & \textbf{Dialogue Length} \\
\midrule
I2EBench \cite{ma2024i2ebench} & \cmark & \cmark & \xmark & \cmark & 1 \\
EditBench \cite{wang2023imagen} & \cmark & \cmark & \xmark & \cmark & 1 \\
EditVal \cite{basu2023editval} & \cmark & \xmark & \xmark & \xmark & 1  \\
EmuEdit \cite{sheynin2024emu} & \cmark & \xmark & \xmark & \xmark & 1 \\
AnyEdit \cite{Yu_2025_CVPR} & \xmark & \cmark & \xmark & \xmark & 1 \\
CompBench \cite{jia2025compbench} & \cmark & \cmark & \xmark & \xmark & 2  \\
MagicBrush \cite{zhang2023magicbrush} & \cmark & \xmark & \xmark & \xmark & 3  \\
ImgEdit-Bench \cite{ye2025imgedit} & \cmark & \xmark & \xmark & \xmark & 3  \\
MuCIE \cite{zhou2025multi} & \xmark & \xmark & \xmark & \xmark & 5  \\
\midrule
\textbf{Omni IIE Bench (Ours)} & \textbf{\cmark} & \textbf{\cmark} & \textbf{\cmark} & \textbf{\cmark} & \textbf{16} \\
\bottomrule
\end{tabular}%
}
\end{table*}

\section{Omni IIE Bench Dataset}
\label{sec:dataset}

Constructing a practical benchmark for diagnosing \textbf{Instruction-based Image Editing (IIE)} models is a challenging engineering task. 
As shown in Figure ~\ref{fig:overview} and Table \ref{tab:compare}, we propose \textbf{Omni IIE Bench}, a benchmark for evaluating practical image editing, rigorously reviewed through manual assessment.

Omni IIE Bench features two core attributes:

\textbf{Diagnostic Power.}
It provides complementary single-turn and multi-turn diagnostic tracks to assess models’ semantic understanding, editing consistency, and stability across iterative modifications.

\textbf{Strict Review.}
Seed images are drawn from 12 public datasets. 
All generated editing instructions and edited images are manually reviewed by professional designers and computer vision researchers to ensure that the data aligns with practical application scenarios.

\subsection{Data Acquisition and Sourcing}
\label{sec:data_acquisition}

To ensure the benchmark's diversity and practical scene coverage, our initial image pool is sampled from 12 public datasets, categorized by content and style into four types:
\begin{itemize}
    \item \textbf{Multimodal Instruction and Conversation:} \texttt{allava-instruct} \cite{chen2024allava}, \texttt{llava-wild} \cite{liu2023visual}, \texttt{sharegpt4o} \cite{chen2025sharegpt}
    \item \textbf{Image Captioning and Description:} \texttt{sharegpt4v} \cite{chen2024sharegpt4v}, \texttt{docci} \cite{onoe2024docci}, \texttt{pixelprose} \cite{singla2024pixels}
    \item \textbf{Traditional Visual Benchmark:} \texttt{coco2017val} \cite{lin2014microsoft}, \texttt{flickr30k} \cite{young2014image}, \texttt{GQA} \cite{hudson2019gqa}
    \item \textbf{Specialized Domain:} \texttt{wikiart} \cite{saleh2015large}, \texttt{SUN397} \cite{xiao2010sun}, \texttt{imageinwords} \cite{garg2024imageinwords}
\end{itemize}

All sampling processes use a fixed random seed $seed = 42$ to ensure reproducibility. We constructed independent seed pools for the two diagnostic tracks. For the \textbf{single-turn consistency seed pool}, we randomly sampled 200 images from each of the 12 datasets, forming a 2,400-image main seed pool. For the \textbf{multi-turn coordination seed pool}, we randomly sampled 58 images from each dataset, forming a 696-image multi-turn dialogue seed pool, used to construct continuous, multi-scale editing dialogues.

\subsection{Data Generation and Annotation Pipeline}
\label{sec:data_pipeline}

To ensure that Omni IIE Bench is both scientifically rigorous and aligned with practical application scenarios, we established detailed annotation type definitions and constructed a three-stage data generation and filtering pipeline.

\subsubsection{Annotation Types}
\label{sec:annotation_types}
All samples are organized as quadruplets $(I_{source}, T_{mod}, I_{gt}, M_{gt})$, where $I_{source}$ is the source image, $T_{mod}$ is the editing instruction, $I_{gt}$ is the target image, and $M_{gt}$ is the ground-truth mask.

\paragraph{Multi-Turn Diagnosis.}
Contains 696 sets of cross-scale multi-turn dialogues. Each set simulates a user's continuous interaction with the model, with instructions dynamically interspersing \textbf{attribute modification} and \textbf{entity replacement}, to evaluate the model's contextual understanding, instruction coordination, and error accumulation capabilities in continuous editing.

\subsubsection{Generation and Annotation Process}
\label{sec:gen_and_annotation_process}

\paragraph{Stage 1: Automated Candidate Data Generation}
This stage utilizes advanced vision-language and generative models to automatically generate candidate data. (1) \textbf{Image Captioning:} We use the \texttt{gpt-4o}\cite{openai2024gpt4ocard} model to generate fine-grained scene descriptions for the 2,400 main seed pool images and the 696 multi-turn dialogue seed pool images, covering target objects, attributes, actions, and environmental relationships. (2) \textbf{Instruction Generation:} The \texttt{gpt-4o} model then generates editing instruction sets based on these scene descriptions. For single-turn consistency diagnosis, it generates dual-task instruction pairs with opposing semantics. For multi-turn coordination, it generates multi-turn instruction sequences that alternate across semantic scales. (3) \textbf{GT Image Generation:} We employ \texttt{Nano Banana} \cite{google2025nanoBanana} to ensure the high fidelity of the ground-truth image generation.

\paragraph{Stage 2: Automated Mask Generation}
The core task of this stage is to automatically generate high-quality ground-truth masks for all generated candidate samples, ensuring spatial alignment with the instruction's intent. (1) \textbf{Instruction Parsing:} Directly inputting the full natural language instruction $T_{mod}$ into \texttt{GroundingDINO} \cite{liu2023grounding} is infeasible, as it requires clear entity nouns. Therefore, we designed an instruction parsing step. We utilize \texttt{gpt-4o} as the instruction parser. The parser's task is to analyze the intent of $T_{mod}$ and extract only the core entity noun being edited. For attribute modification tasks, the parser likewise extracts its carrier; for entity replacement tasks, it extracts the original entity being replaced. (2) \textbf{Mask Extraction:} After instruction parsing, the extracted entity nouns are passed as text prompts to the \texttt{Grounded-SAM} framework \cite{ren2024grounded} for fully automated mask extraction. The process first activates the \texttt{GroundingDINO} module to receive the entity nouns, and its output candidate bounding boxes are passed as spatial prompts to the \texttt{SAM} module \cite{kirillov2023segany}. \texttt{SAM} finally performs high-precision instance segmentation based on these spatial prompts, generating the binarized ground-truth mask $M_{gt}$.

\begin{figure}
    \centering
    \includegraphics[width=1\linewidth]{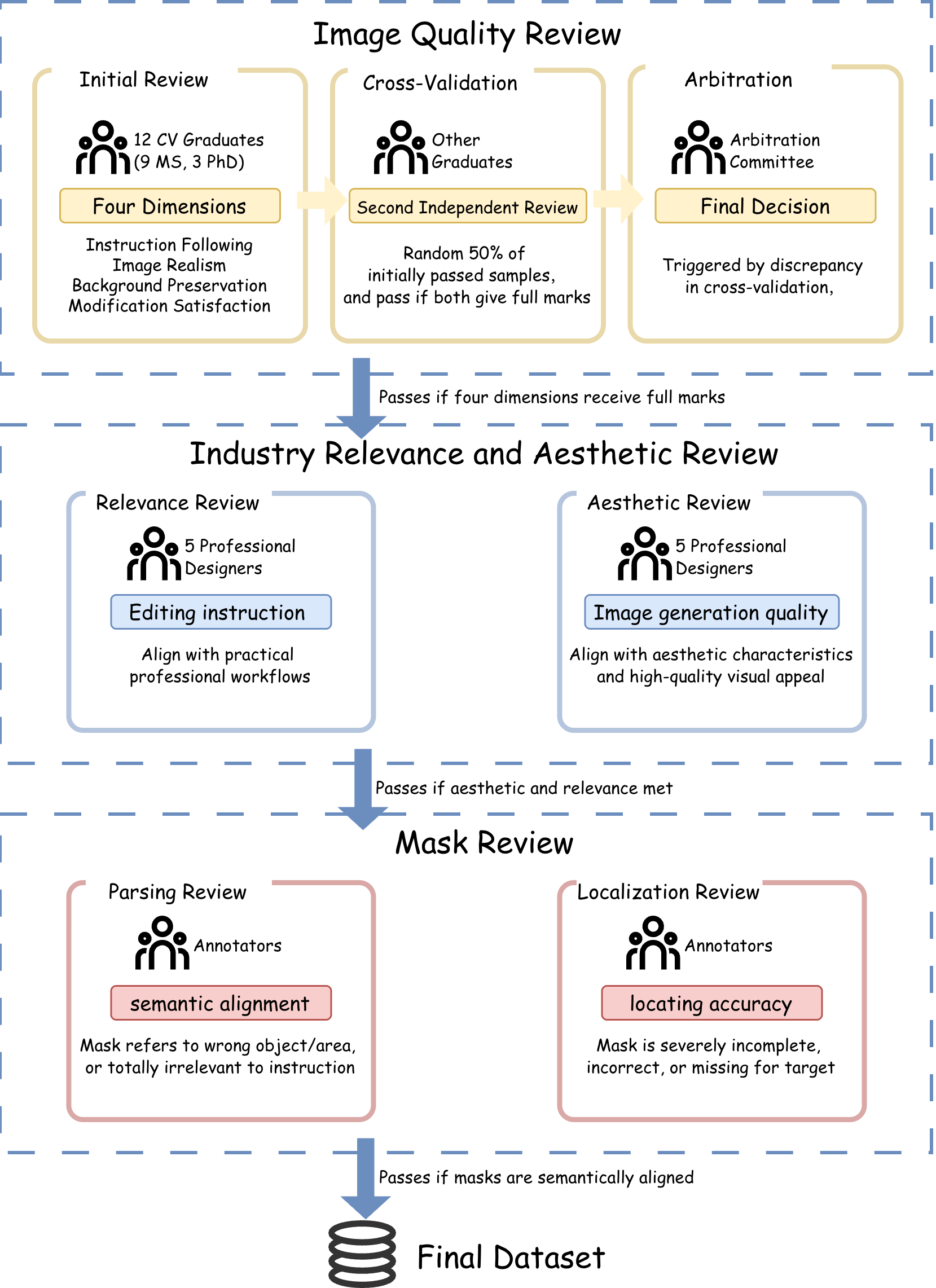}
    \caption{Overview of manual curation and filtering}
    \label{fig:overview of human filtering}
\end{figure}

\vspace{-10pt}

\paragraph{Stage 3: Manual Curation and Filtering}
This stage is the most critical juncture for ensuring the final quality of Omni IIE Bench benchmark. To fully leverage our team's professional capabilities, we designed a hierarchical, multi-pass manual quality assessment process. An overview of the manual curation and filtering process can be found in Figure ~\ref{fig:overview of human filtering}.

\textbf{Annotation Team and Calibration.}
The annotation team includes 12 CV graduate students and 5 AI image editing designers. Before formal annotation, all members undergo a calibration phase to align on detailed guidelines and key metrics such as artifacts, semantic drift, and background pollution, ensuring consistent evaluation standards.
Each candidate sample is manually scored from 1 to 3 on four dimensions: 

\textbf{Image Quality Review.}
Each candidate sample is first manually scored from 1 to 3 on four dimensions:
\begin{itemize}
    \item \textbf{Instruction Following}: Adherence to the editing instruction.
    \item \textbf{Image Realism}: Naturalness and absence of obvious artifacts.
    \item \textbf{Background Preservation}: Integrity of unedited regions.
    \item \textbf{Modification Satisfaction}: Overall acceptability of the editing result.
\end{itemize}
This review follows a rigorous workflow: (1) \textbf{Annotator Initial Review:} 9 master’s and 3 PhD annotators independently score each sample across the four dimensions. (2) \textbf{Cross-Validation and Arbitration:} To ensure quality, 50\% of initially passed samples are randomly selected for a second independent review by another annotator. If both annotators give full marks, the sample passes; if there is a discrepancy, it is submitted to an Arbitration Committee of 5 designers, who make the final decision through discussion and voting.

\textbf{Industry Relevance and Aesthetic Review.}
All samples that passed the initial quality check are then entered into an independent industry relevance and aesthetic review stage. This stage is conducted by a Review Team of 5 professional designers proficient in AI image editing software. The Review Task involves the expert team re-evaluating the image generation quality and the editing instruction $T_{mod}$ itself, assessing whether they align with practical professional workflows and aesthetic characteristics in areas such as advertising design, e-commerce, or film concept art. The Filtering Criteria are based on the expert team's subjective assessment of the instruction’s industry relevance and aesthetic characteristics, and samples deemed not to meet practical work requirements are removed.

\textbf{Mask Review.}
After the automated mask generation, the masks undergo a rapid manual review. This step is not for pixel-level correction; annotators only make a Pass or Fail judgment on the semantic alignment of the mask with the instruction. The sole purpose of this is to filter out samples with severe misalignment caused by parsing errors or \texttt{GroundingDINO} localization failures.

\textbf{Final Filtering Criteria.} We employ an extremely strict dual zero-tolerance filtering standard. A sample must meet both of the following conditions to be accepted into the final Omni IIE Bench benchmark: (1) \textbf{Quality Standard:} Judged to have received full marks on all four dimensions in the Image Quality Review. (2) \textbf{Relevance Standard:} Judged as Relevant by the professional designer team in the Industry Relevance and Aesthetic Review. This rigorous dual standard, combined with the mask review, ensures every sample is high-quality, relevant, and spatially valid. Ultimately, we were left with 2856 images.

\subsection{Data Statistics}
\label{sec:data_statistics}

\subsubsection{Curation Pipeline and Source Distribution}
\label{sec:stats_funnel}

\textbf{Curation Pipeline.}
We employed a rigorous, zero-tolerance filtering standard across both diagnostic tracks. For the single-turn track, from 4,800 initial candidate samples, only 50.86\% passed the multi-dimensional quality review. Of those, an additional 29.33\% were subsequently discarded by industry experts for lacking practical relevance, resulting in a final acceptance rate of just 35.94\%, yielding 1,725 samples. For the multi-turn track, from 696 initial dialogue groups containing 5,421 images, a similar filtering process yielded 260 final groups, which represents a 37.36\% acceptance rate for groups, and 1,131 final images, representing a 20.86\% acceptance rate for images. This multi-stage process, with a combined final acceptance of 2,856 images from 10,221 initial candidates, ensures the reliability and practical value of the final Omni IIE Bench.

\begin{figure}
    \centering
    \includegraphics[height=0.6\linewidth]{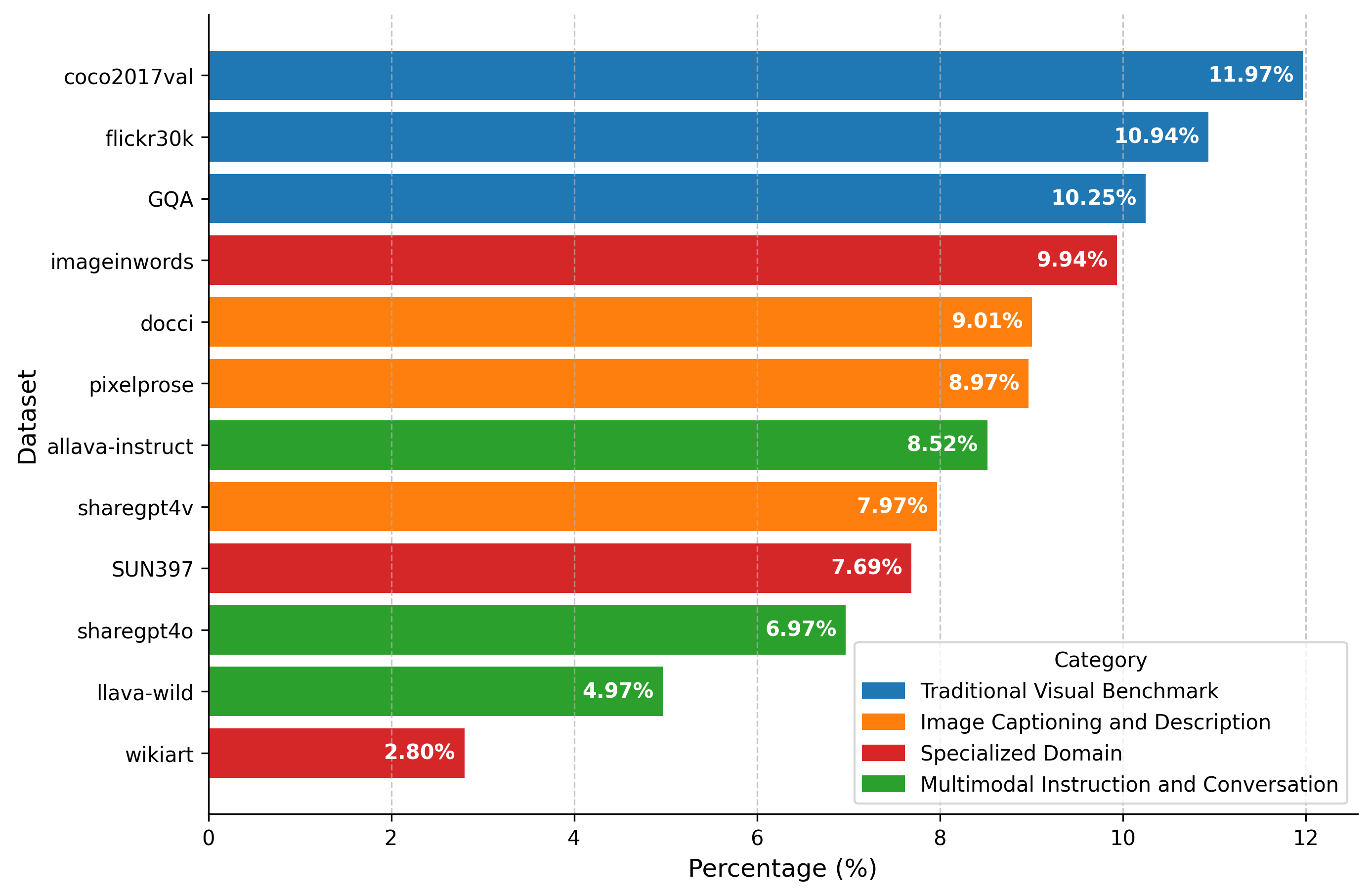}
    \caption{Source Distribution of Datasets in Omni IIE Bench}
    \label{fig:source_distribution}
\end{figure}

\textbf{Source Distribution.}
As shown in Figure \ref{fig:source_distribution}, our benchmark is sourced from 12 distinct datasets. The distribution is well-balanced, with the largest single source accounting for only 11.97\% of the data. This balanced mix ensures our evaluation is not biased towards any single data type and robustly covers diverse scenarios.

\subsubsection{Diagnostic Attribute Analysis}
\label{sec:stats_diagnostic}

\textbf{Mask Attributes.}
Mask area statistics validate our division of semantic scales. We find that attribute modification tasks' mask areas tend to be concentrated in a small range, while entity replacement tasks' mask areas are more broadly distributed.

\textbf{Multi-Turn Coordination Attributes.}
The multi-turn track contains 260 dialogue groups with an average depth of 4.35 turns. Within the 1,131 total editing turns, we recorded 322 attribute-to-entity and 178 entity-to-attribute scale switches, ensuring thorough testing of dynamic, cross-scale coordination.

\begin{table*}[h]
\centering
\setlength{\tabcolsep}{10pt}
\caption{
Performance comparison of IIE models on a single turn. Text colors indicate top performances: 
\textcolor{orange!90!black}{1st}, 
\textcolor{gray!70!black}{2nd}, 
\textcolor{brown!70!black}{3rd}. 
Arrows ($\uparrow$ / $\downarrow$) indicate whether higher or lower values are better.
The Overall score is calculated by the formula $\frac{1}{4} \left[ \frac{3 - (\sum \text{LPIPS})}{3} + \frac{\sum \text{CLIP}}{3} + \text{QA} + \text{SSIM} \right]$, where $\Sigma\text{LPIPS}$ and $\Sigma\text{CLIP}$ are the sums of their respective FG, BG, and ALL columns.
}
\label{tab:main single turn}
\renewcommand{\arraystretch}{0.8}

\resizebox{\textwidth}{!}{
\begin{tabular}{l|ccc|ccc|c|cc|c}
\toprule
\multirow{2}{*}{Model} 
& \multicolumn{3}{c|}{LPIPS~($\downarrow$)} 
& \multicolumn{3}{c|}{CLIP~($\uparrow$)} 
& \multirow{2}{*}{QA~($\uparrow$)} 
& \multicolumn{2}{c|}{Image Quality~($\uparrow$)} 
& \multirow{2}{*}{Overall} \\ 
\cmidrule(lr){2-4} \cmidrule(lr){5-7} \cmidrule(lr){9-10}
& FG & BG & ALL
& FG & BG & ALL
& 
& PSNR & SSIM 
& \\
\midrule
Qwen-image-edit\cite{wu2025qwenimagetechnicalreport} & \textcolor{gray!70!black}{0.245} & 0.327 & 0.450 & \textcolor{orange!90!black}{0.887} & \textcolor{gray!70!black}{0.891} & \textcolor{gray!70!black}{0.889} &  \textcolor{orange!90!black}{0.744} & 14.401 & 0.455 & \textcolor{orange!90!black}{0.687} \\
InstructPix2Pix\cite{brooks2023instructpix2pix} & 0.284 & 0.441 & 0.569 & 0.774 & 0.775 & 0.841 & 0.316 & 12.423 & 0.438 & 0.530 \\
ICEdit\cite{zhang2025context} & \textcolor{brown!70!black}{0.252} & \textcolor{brown!70!black}{0.295} & \textcolor{gray!70!black}{0.425} & \textcolor{gray!90!black}{0.867} & \textcolor{brown!70!black}{0.875} & \textcolor{brown!70!black}{0.868} & 0.453 & \textcolor{gray!70!black}{15.541} & \textcolor{gray!70!black}{0.507} & \textcolor{brown!70!black}{0.626} \\
MGIE\cite{fu2024mgie} & 0.261 & \textcolor{gray!70!black}{0.294} & \textcolor{brown!70!black}{0.426} & 0.858 & 0.864 & 0.859 & 0.070 & \textcolor{brown!70!black}{14.733} & \textcolor{brown!70!black}{0.480} & 0.520 \\
HIVE\cite{zhang2024hive} & 0.287 & 0.406 & 0.526 & 0.849 & 0.796 & 0.794 & 0.259 & 13.423 & 0.414 & 0.527 \\
FLUX\cite{batifol2025flux} & 0.283 & 0.428 & 0.552 & \textcolor{brown!70!black}{0.863} & 0.871 & 0.868 & \textcolor{gray!70!black}{0.636} & 12.553 & 0.375 & 0.614 \\
HQEdit\cite{hui2024hq} & 0.327 & 0.557 & 0.689 & 0.794 & 0.691 & 0.694 & 0.322 & 9.259 & 0.304 & 0.457 \\
Step1X\cite{liu2025step1x-edit} & \textcolor{orange!90!black}{0.230} & \textcolor{orange!70!black}{0.259} & \textcolor{orange!70!black}{0.379} & \textcolor{orange!90!black}{0.887} & \textcolor{orange!90!black}{0.903} & \textcolor{orange!90!black}{0.899} & \textcolor{brown!70!black}{0.580} & \textcolor{orange!90!black}{15.845} & \textcolor{orange!90!black}{0.533} & \textcolor{gray!90!black}{0.680} \\
\bottomrule
\end{tabular}
}

\end{table*}

\begin{table*}[h]
\centering
\setlength{\tabcolsep}{10pt}
\caption{
Performance comparison of IIE models on multi turns. Text colors indicate top performances: 
\textcolor{orange!90!black}{1st}, 
\textcolor{gray!70!black}{2nd}, 
\textcolor{brown!70!black}{3rd}. 
Arrows ($\uparrow$ / $\downarrow$) indicate whether higher or lower values are better.}
\label{tab:main multi turn}
\renewcommand{\arraystretch}{0.8}

\resizebox{\textwidth}{!}{
\begin{tabular}{l|ccc|ccc|c|cc|c}
\toprule
\multirow{2}{*}{Model} 
& \multicolumn{3}{c|}{LPIPS~($\downarrow$)} 
& \multicolumn{3}{c|}{CLIP~($\uparrow$)} 
& \multirow{2}{*}{QA~($\uparrow$)} 
& \multicolumn{2}{c|}{Image Quality~($\uparrow$)} 
& \multirow{2}{*}{Overall} \\ 
\cmidrule(lr){2-4} \cmidrule(lr){5-7} \cmidrule(lr){9-10}
& FG & BG & ALL
& FG & BG & ALL
& 
& PSNR & SSIM 
& \\
\midrule
Qwen-image-edit & \textcolor{gray!70!black}{0.254} & \textcolor{brown!90!black}{0.390} & \textcolor{gray!70!black}{0.510} & \textcolor{gray!70!black}{0.888} & \textcolor{gray!70!black}{0.879} & \textcolor{gray!70!black}{0.877} & \textcolor{orange!90!black}{0.818} & \textcolor{brown!90!black}{13.521} & 0.391 & \textcolor{orange!90!black}{0.676} \\
InstructPix2Pix & 0.296 & 0.552 & 0.669 & 0.842 & 0.701 & 0.703 & 0.560 & 10.910 & \textcolor{brown!90!black}{0.393} & 0.549 \\
ICEdit & \textcolor{brown!90!black}{0.263} & \textcolor{gray!70!black}{0.388} & \textcolor{brown!90!black}{0.515} & \textcolor{brown!90!black}{0.875} & 0.845 & 0.842 & 0.524 & \textcolor{gray!70!black}{14.537} & \textcolor{gray!70!black}{0.435} & \textcolor{brown!90!black}{0.606} \\
MGIE & 0.325 & 0.580 & 0.685 & 0.801 & 0.710 & 0.723 & 0.054 & 10.508 & 0.350 & 0.404\\
HIVE & 0.294 & 0.484 & 0.607 & 0.847 & 0.771 & 0.770 & 0.487 & 12.415 & 0.337 & 0.540 \\
FLUX & 0.275 & 0.478 & 0.599 & 0.867 & \textcolor{brown!90!black}{0.860} & \textcolor{brown!90!black}{0.857} & \textcolor{gray!70!black}{0.678} & 11.903 & 0.332 & 0.605\\
HQEdit & 0.317 & 0.590 & 0.710 & 0.819 & 0.677 & 0.682 & 0.524 & 9.010 & 0.292 & 0.501 \\
Step1X & \textcolor{orange!90!black}{0.250} & \textcolor{orange!90!black}{0.344} & \textcolor{orange!90!black}{0.453} & \textcolor{orange!90!black}{0.896} & \textcolor{orange!90!black}{0.888} & \textcolor{orange!90!black}{0.885} & \textcolor{brown!90!black}{0.610} & \textcolor{orange!90!black}{14.837} & \textcolor{orange!90!black}{0.464} & \textcolor{gray!70!black}{0.654} \\
\bottomrule
\end{tabular}
}

\end{table*}

\section{Evaluation Methodology}
\label{sec:evaluation}

To fairly and comprehensively evaluate IIE models, we propose a decoupled diagnostic evaluation framework. Our methodology decouples the evaluation into three core dimensions: (1) \textbf{Global Image Quality}, (2) \textbf{Decoupled Regional Fidelity}, and (3) \textbf{Instruction Compliance}. This section will detail our evaluation pipeline and all metrics used.

\subsection{Evaluation Pipeline and Preprocessing}
\label{subsec:pipeline}

Our evaluation pipeline loads samples from a JSON manifest, which provides paths to the source image $I_{source}$, the ground-truth target image $I_{gt}$, and the ground-truth mask $M_{gt}$.
To ensure all comparisons are conducted on a uniform scale, we define a strict preprocessing step. All images are resampled to a standard resolution of (768, 768) using lanczos interpolation. All masks are similarly resampled to (768, 768), but using nearest interpolation to preserve the sharpness of their binary boundaries.
We define the foreground mask $M_{fg} = M_{gt}$ and the background mask $M_{bg} = 1 - M_{gt}$. All decoupled metrics are computed on these specific regions.

\subsection{Global Image Quality Metrics}
\label{subsec:global_metrics}
This dimension assesses the overall quality of $I_{gen}$ compared to $I_{gt}$. We employ four standard metrics: 
\textbf{PSNR} and \textbf{SSIM} \cite{wang2004image} evaluate pixel-level fidelity and structural similarity.
\textbf{LPIPS} \cite{zhang2018unreasonable} measures perceptual similarity. It is computed as a weighted distance between deep features $F$ from network layers $l$, where $H_l$ and $W_l$ are the feature map dimensions, $\sum_{h,w}$ iterates over all spatial positions, and $\odot$ denotes element-wise multiplication.
{
\setlength{\abovedisplayskip}{3pt}
\begin{multline}
\text{LPIPS}(x, x_0) = \\
\sum_l \frac{1}{H_l W_l} \sum_{h,w}
\| w_l \odot (\hat{F}^l_{hw}(x) - \hat{F}^l_{hw}(x_0)) \|_2^2,
\end{multline}
}
\textbf{CLIP Score} \cite{radford2021learning} measures global semantic consistency. It is defined as the cosine similarity between the CLIP image embeddings ($E_I$).
\begin{equation}
\text{CLIP}(I_1, I_2) = \frac{E_I(I_1) \cdot E_I(I_2)}{||E_I(I_1)|| \cdot ||E_I(I_2)||} 
\end{equation}

\subsection{Decoupled Regional Fidelity Metrics}
\label{subsec:decoupled_metrics}
Global metrics can average out local errors. We therefore decouple the evaluation into foreground and background regions using the ground-truth mask $M_{fg} = M_{gt}$ and its inverse $M_{bg} = 1 - M_{gt}$.

\subsubsection{Editing Fidelity}
This evaluation compares $I_{gen}$ and $I_{gt}$ within the $M_{fg}$ region. To compute foreground metrics, we first generate masked images $I_{gen}^{fg}$ and $I_{gt}^{fg}$ by zeroing out pixels outside $M_{fg}$. We then compute \textbf{FG-LPIPS} to measure perceptual fidelity and the \textbf{FG-CLIP Score} for semantic fidelity.
\begin{equation}
\text{FG-LPIPS} = \text{LPIPS}(I_{gen}^{fg}, I_{gt}^{fg})
\end{equation}
\begin{equation}
\text{FG-CLIP} = \text{CLIP}(I_{gen}^{fg}, I_{gt}^{fg})
\end{equation}

\subsubsection{Background Fidelity}
This evaluation quantifies the consistency of the $M_{bg}$ region. Symmetrically, we create $I_{gen}^{bg}$ and $I_{gt}^{bg}$ by setting the foreground pixels to 0.
We then compute \textbf{BG-LPIPS} and the \textbf{BG-CLIP Score} between these two background-only images.
\begin{equation}
    \text{BG-LPIPS} = \text{LPIPS}(I_{gen}^{bg}, I_{gt}^{bg})
\end{equation}
\begin{equation}
    \text{BG-CLIP} = \text{CLIP}(I_{gen}^{bg}, I_{gt}^{bg})
\end{equation}

\begin{figure*}
    \centering
    \includegraphics[width=0.88\linewidth]{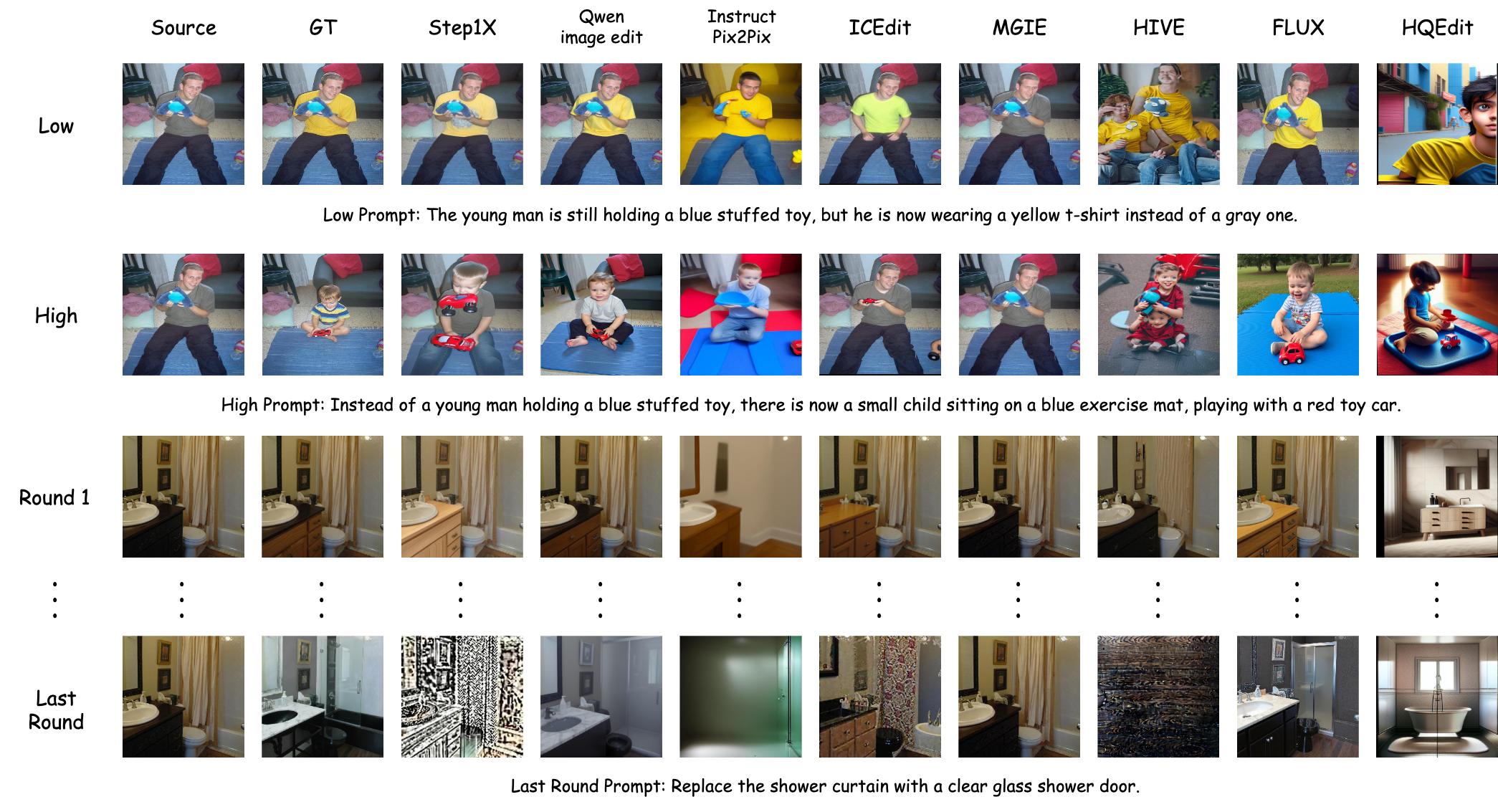}
    \caption{Visualizing Single-Turn and Multi-Turn Consistency Across Multiple Models}
    \label{fig:sample of figure}
\end{figure*}
\section{Experiment}
\label{sec:experiment}

\subsection{Instruction Compliance Metrics}
\label{subsec:compliance_metrics}

The above metrics measure image-to-image fidelity but cannot assess whether $I_{gen}$ semantically follows the instruction $T_{mod}$. To address this, we introduce an evaluation that directly links the text instruction with the generated image using automated QA generation and an MLLM referee.\\
\textbf{Step 1: Automated QA Pair Generation.} For each instruction $T_{mod}$, we automatically generate 1–3 verifiable QA pairs using \texttt{gpt-4o}, providing a ground-truth answer.\\
\textbf{Step 2: MLLM Referee.} We input $I_{gen}$, the question, and ground-truth answer into \texttt{gpt-4o} to perform multimodal verification, asking it to return True or False on whether the answer is semantically correct for the image.\\
\textbf{Step 3: Scoring.} We use a strict one-strike policy: a sample scores 1.0 only if all its QA pairs are True; if any are False, the score is 0.0, ensuring only fully compliant edits get a high score.

In this section, we conduct a comprehensive evaluation of 8 mainstream Instruction-based Image Editing (IIE) models using Omni IIE Bench benchmark.

\subsection{Experimental Setup}
\label{subsec:exp_setup}

We selected 8 representative IIE models for evaluation, covering the diverse technical approaches currently available. These include Qwen-image-edit, MGIE, InstructPix2Pix, ICEdit, HIVE, hqedit, FLUX, and Step1X.

\paragraph{Evaluation Metrics.} We employ the decoupled evaluation framework defined in Section~\ref{sec:evaluation}, which includes global image quality metrics, regional fidelity metrics, and instruction compliance metrics.

\subsection{Single-turn Consistency Diagnosis}
\label{subsec:exp_single_turn}

\begin{figure*}
    \centering
    \includegraphics[width=0.9\linewidth]{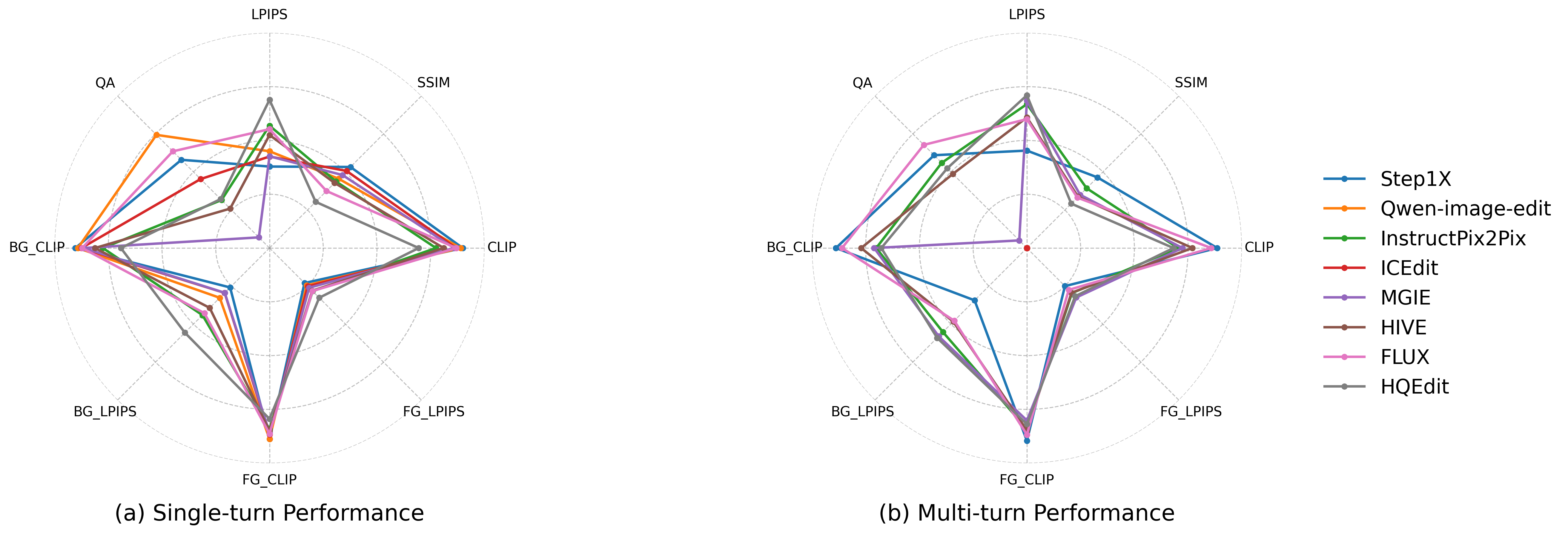}
    \caption{Comparison of 8 IIE models across 8 metrics on Omni IIE Bench. The charts show normalized scores for (a) the Single-turn Consistency task and (b) the Multi-turn Coordination task.}
    \label{fig:radar}
\end{figure*}

\begin{figure}
    \centering
    \includegraphics[height=0.35\linewidth]{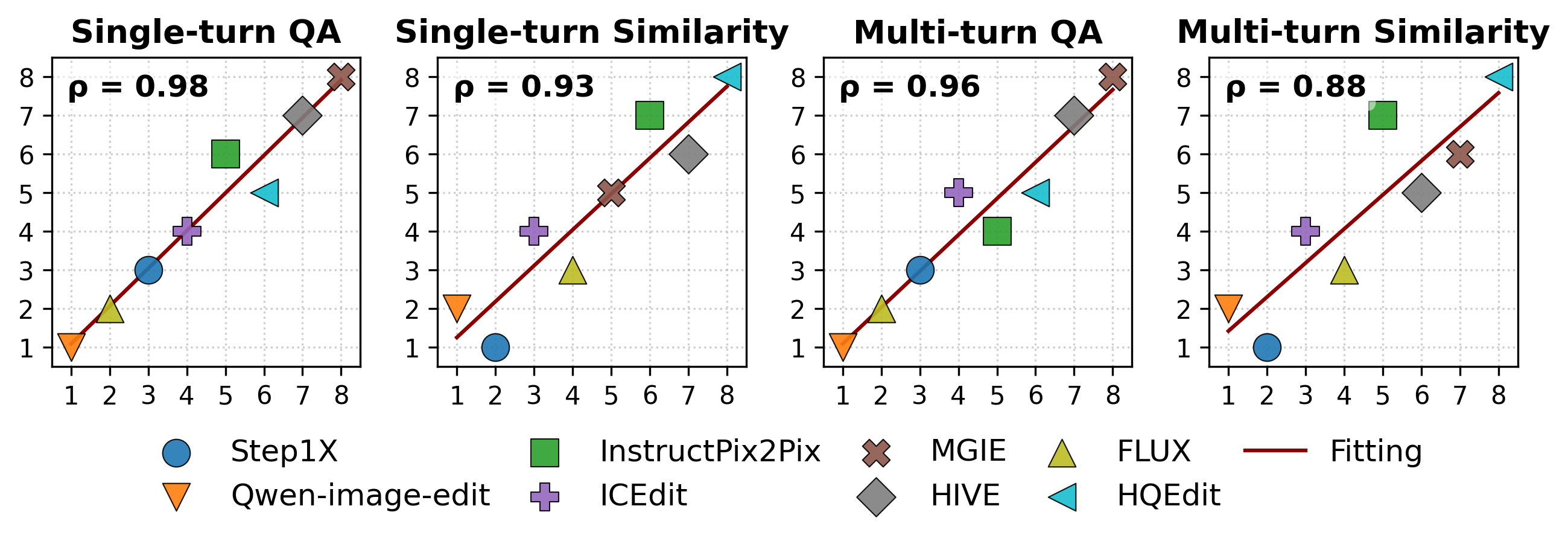}
    \caption{Alignment between Omni IIE Bench ranks (Y-axis) and human evaluation ranks (X-axis).}
    \label{fig:human correlation}
\end{figure}

\paragraph{Overall Evaluation Results.}
As shown in Table ~\ref{tab:main single turn}, Omni IIE Bench reveals significant differences among the models. Qwen-image-edit performs the best, while HQEdit performs the worst. There is a considerable gap between instruction-following ability and image generation quality. For example, although MGIE achieves high image generation quality, its instruction-following capability is very poor. This allows it to score highly on many benchmarks that do not evaluate instruction compliance, whereas Omni IIE Bench is able to detect this issue. As shown in the examples in Figure \ref{fig:sample of figure}, Qwen-Image-Edit outperforms the other models, whereas MGIE outputs are almost identical to the original images, demonstrating extremely poor instruction-following ability.

\paragraph{Effect of Semantic Scale}
Regarding the impact of instruction-editing semantic scales on image editing models, we provide a detailed analysis in Appendix Section \ref{sec:metric detail}. Please refer to the appendix for full details.

\subsection{Multi-turn Coordination Diagnosis}
\label{subsec:exp_multi_turn}
\paragraph{Overall Evaluation Results.} 
Unlike single-turn dialogues, as shown in Table \ref{tab:main multi turn}, multi-turn generation performance drops for all models due to error accumulation, though Qwen-Image-Edit and Step1X remain relatively strong, while MGIE declines further.

\paragraph{Instruction-following Ability}
For multi-turn dialogues, to better simulate practical scenarios, the instruction length for each turn is generally shorter, resulting in a significantly lower difficulty for QA generation compared to single-turn dialogues. Consequently, the instruction-following evaluation is inherently easier in this setting. As a result, for the same model, the QA scores in Table \ref{tab:main multi turn} are often higher than those in Table \ref{tab:main single turn}.

\paragraph{Background Preservation Capability.}
Comparing Tables \ref{tab:main single turn} and \ref{tab:main multi turn}, multi-turn dialogues clearly cause a substantial drop in background preservation across all models, primarily due to error accumulation. In real-world scenarios, where designers frequently perform multiple edits, such errors accumulate continuously, highlighting a key limitation of current mainstream models, according to Omni IIE Bench.

\subsection{Human Alignment}
\label{subsec:human alignment}
To validate that the evaluation metrics are accurate and representative, we conducted a human alignment assessment experiment. In this experiment, we evaluated the alignment between the metrics and human judgment from two aspects: image similarity and instruction compliance. For instruction compliance, we focused on the ranking according to the model’s QA metric. For image similarity, since humans can only subjectively assess the similarity between two images, we focused on the ranking based on the model’s CLIP metric.

We selected four annotators for this study, including two PhD students in computer science and two professional designers. From the single-turn results generated by each model, we randomly sampled 100 sets, each consisting of the original image, the image generated by \textit{Nano Banana}, and the corresponding instruction. Similarly, from the multi-turn results, we randomly sampled 20 complete multi-turn dialogues, each including the original image, the Nano Banana-generated images, and the corresponding instructions. The four annotators were then asked to score instruction compliance and similarity to the ground-truth image, using a scale from 1 to 3. The mean score for each sample was calculated, and final rankings were determined based on these mean scores.

We plot the rankings of different models according to Omni IIE Bench against the human rankings in a 2D plot, where the X-axis represents human rankings and the Y-axis represents model rankings. We then compute the correlation coefficient across the corresponding points for each image.

As shown in Figure \ref{fig:human correlation}, the rankings of all metrics exhibit a very high correlation with human subjective rankings, with correlation coefficients all above 0.85. This data demonstrates the accuracy and representativeness of Omni IIE Bench evaluation process.

\section{Conclusion}
\label{sec:conclusion}

This paper addresses a key unresolved issue in current Instruction-based Image Editing (IIE) evaluations: a significant gap exists between benchmark performance and real-world applicability. To tackle this problem, we propose \textbf{Omni IIE Bench} — a high-quality, human-annotated diagnostic benchmark designed to assess the practical performance of IIE models.

Omni IIE Bench includes both single turn and multi turn dialogue evaluations, combined with multi-stage human filtering to achieve a comprehensive assessment of model capabilities. We evaluated 8 mainstream IIE models, and the results show that even state-of-the-art models experience significant drops in image fidelity, semantic accuracy, and background preservation when transitioning from low- to high-semantic-scale tasks. Furthermore, in multi-turn dialogues, error accumulation leads to a substantial decline in performance across nearly all models. Omni IIE Bench provides diagnostic tools and insights to guide the development of more practical IIE models.

{
    \small
    \bibliographystyle{ieeenat_fullname}
    \bibliography{main}
}

\appendix
\onecolumn   
\begin{center}
\Large
\textbf{Appendix}
\end{center}

\section{Cases of Omni IIE Bench}
\begin{figure*}[b]
    \centering
    \includegraphics[width=0.9\linewidth]{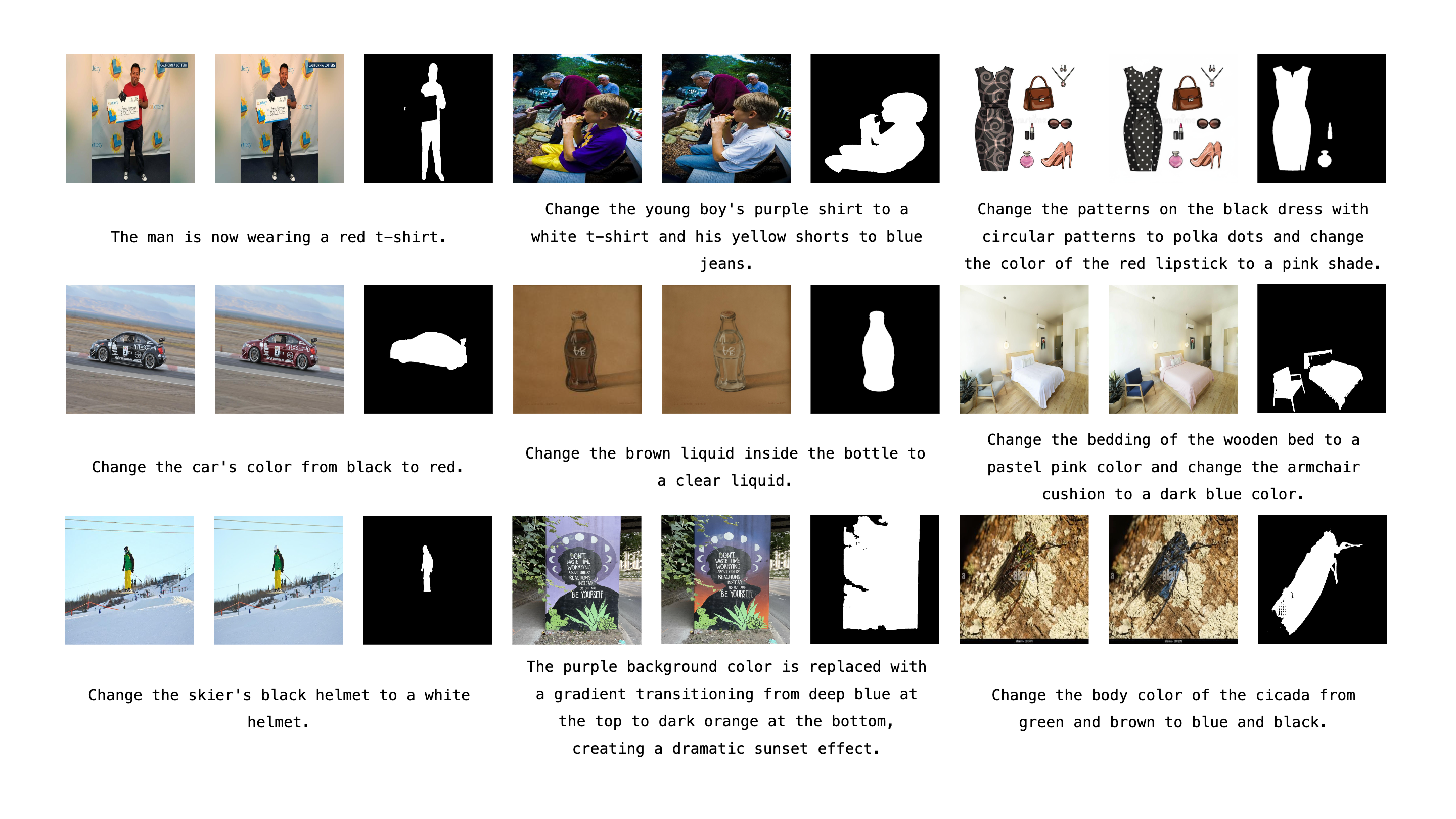}
    \caption{Cases of Single Turn Low}
    \label{fig:single case}
\end{figure*}

\begin{figure*}[b]
    \centering
    \includegraphics[width=0.9\linewidth]{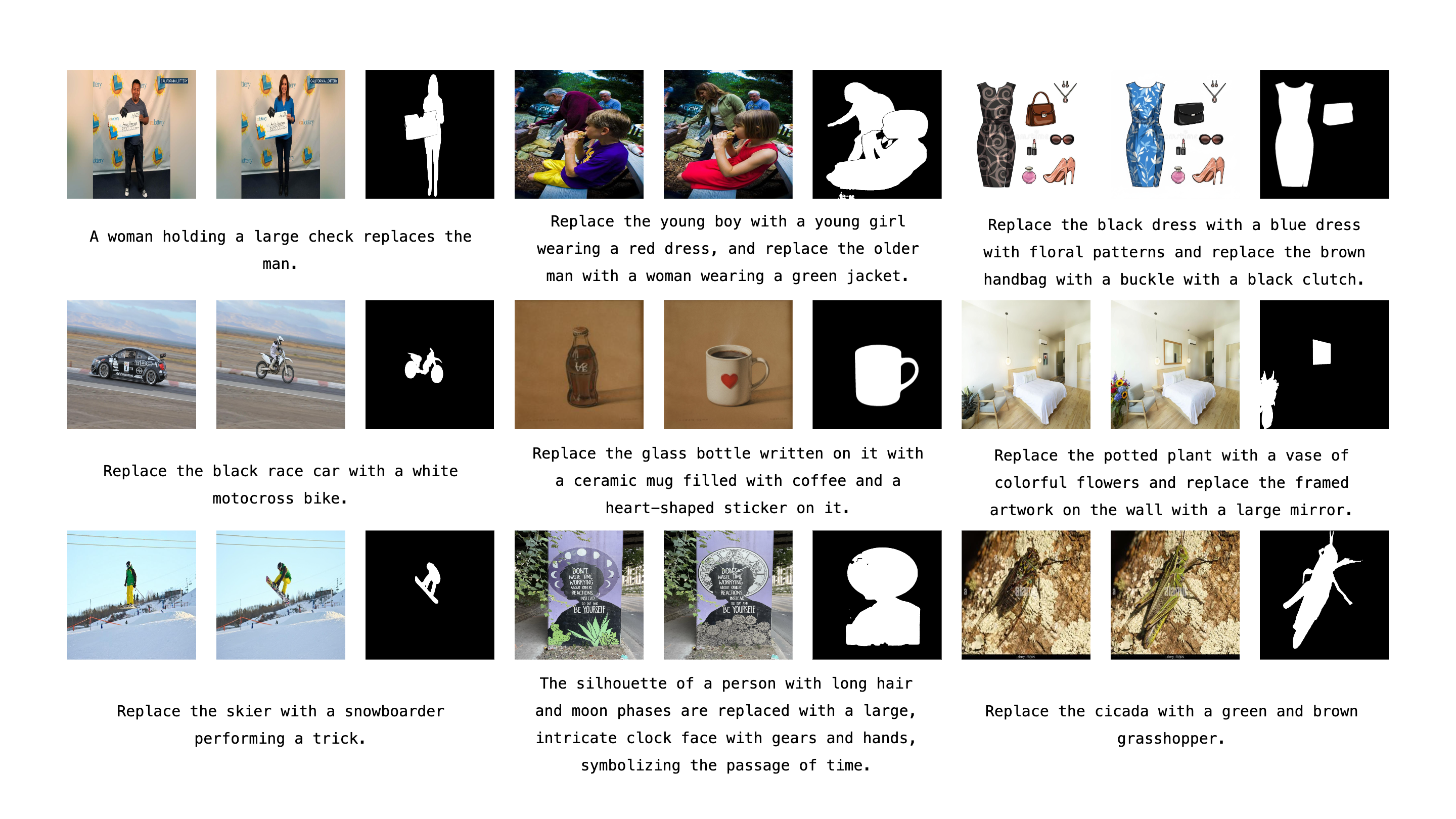}
    \caption{Cases of Single Turn High}
    \label{fig:multi case}
\end{figure*}

This section presents several GT images and their corresponding instructions for single-turn dialogues, including examples of edits at both low and high semantic scales as shown in Figures \ref{fig:single case} and \ref{fig:multi case}.
\FloatBarrier 

\begin{figure*}[b]
    \centering
    \includegraphics[width=0.9\linewidth]{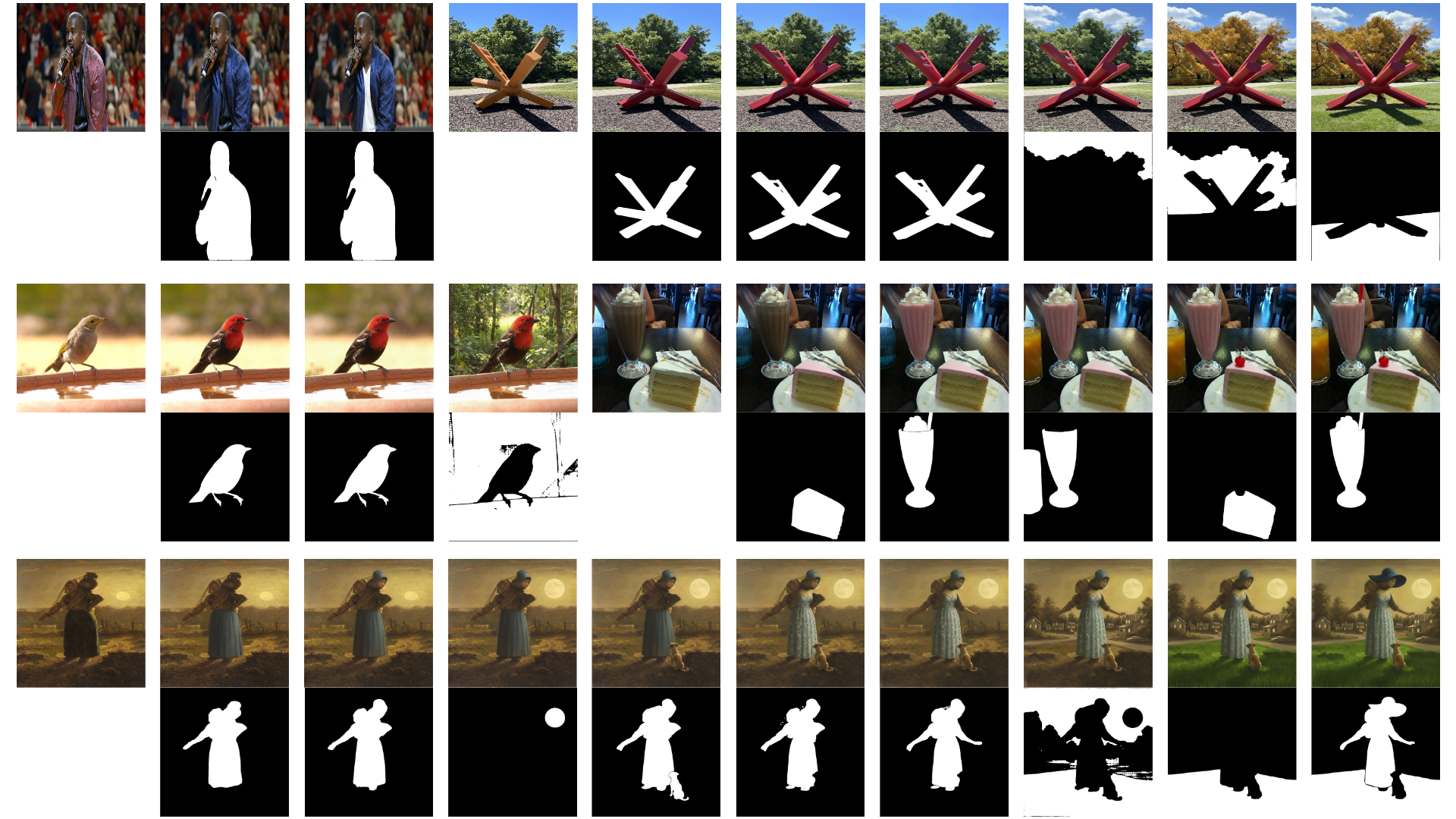}
    \caption{Cases of Multi Turn}
    \label{fig:placeholder}
\end{figure*}

\FloatBarrier

\section{Single-Turn Dialogue Metric Details}
\label{sec:metric detail}
This section presents the metric details of different models at low and high semantic scales during single-turn dialogues.
Overall, regardless of the model, edits at a high semantic scale perform worse than those at a low semantic scale. The specific details are shown in Tables \ref{tab:main_low_grouped} and \ref{tab:main_high_grouped}.

\begin{table*}[h]
\centering
\small
\caption{
Performance comparison of IIE models on the Low group (metrics grouped by category).
Text colors indicate top performances:
\textcolor{orange!90!black}{1st},
\textcolor{gray!70!black}{2nd},
\textcolor{brown!70!black}{3rd}.
Arrows ($\uparrow$ / $\downarrow$) indicate whether higher or lower values are better.
The Overall score is calculated by the formula $\frac{1}{4} \left[ \frac{3 - (\sum \text{LPIPS})}{3} + \frac{\sum \text{CLIP}}{3} + \text{QA} + \text{SSIM} \right]$, where $\Sigma\text{LPIPS}$ and $\Sigma\text{CLIP}$ are the sums of their respective FG, BG, and ALL columns.
}
\label{tab:main_low_grouped}
\renewcommand{\arraystretch}{0.8}
\resizebox{\textwidth}{!}{
\begin{tabular}{l|ccc|ccc|c|cc|c}
\toprule
\multirow{2}{*}{Model} 
& \multicolumn{3}{c|}{LPIPS~($\downarrow$)} 
& \multicolumn{3}{c|}{CLIP~($\uparrow$)} 
& \multirow{2}{*}{QA~($\uparrow$)} 
& \multicolumn{2}{c|}{Image Quality~($\uparrow$)} 
& \multirow{2}{*}{Overall} \\ 
\cmidrule(lr){2-4} \cmidrule(lr){5-7} \cmidrule(lr){9-10}
& FG & BG & ALL
& FG & BG & ALL
& 
& PSNR & SSIM
& \\
\midrule
Qwen-image-edit & \textcolor{gray!70!black}{0.231} & \textcolor{brown!70!black}{0.299} & 0.414 & \textcolor{gray!70!black}{0.910} & \textcolor{gray!70!black}{0.899} & \textcolor{gray!70!black}{0.910} & \textcolor{orange!90!black}{0.780} & 15.323 & 0.474 & \textcolor{gray!70!black}{0.710} \\ 
Instruct\_Pix2Pix & 0.284 & 0.435 & 0.555 & 0.861 & 0.797 & 0.796 & 0.400 & 12.476 & 0.443 & 0.559 \\
ICEdit & \textcolor{brown!70!black}{0.250} & 0.311 & \textcolor{brown!70!black}{0.407} & {0.888} & 0.896 & 0.889 & 0.550 & \textcolor{gray!70!black}{16.107} & \textcolor{gray!70!black}{0.519} & \textcolor{brown!70!black}{0.662} \\ 
MGIE & 0.264 & \textcolor{gray!70!black}{0.294} & \textcolor{gray!70!black}{0.406} & \textcolor{brown!70!black}{0.894} & \textcolor{brown!70!black}{0.898} & \textcolor{brown!70!black}{0.895} & 0.080 & \textcolor{brown!70!black}{15.328} & \textcolor{brown!70!black}{0.495} & 0.539 \\
HIVE & 0.283 & 0.376 & 0.505 & 0.873 & 0.836 & 0.836 & 0.300 & 14.257 & 0.444 & 0.551 \\
FLUX & 0.287 & 0.422 & 0.545 & 0.878 & 0.891 & 0.888 & \textcolor{gray!70!black}{0.720} & 12.769 & 0.376 & 0.641 \\
HQEdit & 0.314 & 0.539 & 0.671 & 0.810 & 0.711 & 0.713 & 0.420 & 9.504 & 0.314 & 0.491 \\
Step1X & \textcolor{orange!90!black}{0.218} & \textcolor{orange!90!black}{0.244} & \textcolor{orange!90!black}{0.353} & \textcolor{orange!90!black}{0.912} & \textcolor{orange!90!black}{0.923} & \textcolor{orange!90!black}{0.920} & \textcolor{brown!70!black}{0.650} & \textcolor{orange!90!black}{16.663} & \textcolor{orange!90!black}{0.547} & \textcolor{orange!90!black}{0.711} \\
\bottomrule
\end{tabular}}
\end{table*}


\begin{table*}[h]
\centering
\small
\caption{
Performance comparison of IIE models on the High group (metrics grouped by category).  
Text colors indicate top performances: 
\textcolor{orange!90!black}{1st}, 
\textcolor{gray!70!black}{2nd}, 
\textcolor{brown!70!black}{3rd}.  
Arrows ($\uparrow$ / $\downarrow$) indicate whether higher or lower values are better.
Arrows ($\uparrow$ / $\downarrow$) indicate whether higher or lower values are better.
The Overall score is calculated by the formula $\frac{1}{4} \left[ \frac{3 - (\sum \text{LPIPS})}{3} + \frac{\sum \text{CLIP}}{3} + \text{QA} + \text{SSIM} \right]$, where $\Sigma\text{LPIPS}$ and $\Sigma\text{CLIP}$ are the sums of their respective FG, BG, and ALL columns.
}
\label{tab:main_high_grouped}
\renewcommand{\arraystretch}{0.8}
\resizebox{\textwidth}{!}{
\begin{tabular}{l|ccc|ccc|c|cc|c}
\toprule
\multirow{2}{*}{Model} 
& \multicolumn{3}{c|}{LPIPS~($\downarrow$)} 
& \multicolumn{3}{c|}{CLIP~($\uparrow$)} 
& \multirow{2}{*}{QA~($\uparrow$)} 
& \multicolumn{2}{c|}{Image Quality~($\uparrow$)} 
& \multirow{2}{*}{Overall} \\ 
\cmidrule(lr){2-4} \cmidrule(lr){5-7} \cmidrule(lr){9-10}
& FG & BG & ALL
& FG & BG & ALL
& 
& PSNR & SSIM
& \\
\midrule
Qwen-image-edit & \textcolor{brown!70!black}{0.264} & 0.360 & 0.495 & \textcolor{orange!90!black}{0.859} & \textcolor{gray!70!black}{0.866} & \textcolor{gray!70!black}{0.866} & \textcolor{orange!90!black}{0.710} & 13.241 & 0.431 & \textcolor{orange!90!black}{0.657} \\
Instruct\_Pix2Pix & 0.283 & 0.448 & 0.586 & 0.816 & 0.749 & 0.747 & 0.230 & 12.356 & 0.431 & 0.498 \\
ICEdit & \textcolor{gray!70!black}{0.254} & \textcolor{gray!70!black}{0.311} & \textcolor{gray!70!black}{0.448} & 0.840 & \textcolor{brown!70!black}{0.849} & \textcolor{brown!70!black}{0.842} & 0.360 & \textcolor{gray!70!black}{14.830} & \textcolor{gray!70!black}{0.492} & \textcolor{brown!70!black}{0.589} \\
MGIE & \textcolor{brown!70!black}{0.264} & \textcolor{brown!70!black}{0.313} & \textcolor{brown!70!black}{0.452} & 0.813 & 0.821 & 0.813 & 0.060 & \textcolor{brown!70!black}{13.983} & \textcolor{brown!70!black}{0.461} & 0.499 \\
HIVE & 0.278 & 0.415 & 0.552 & 0.819 & 0.783 & 0.778 & 0.220 & 13.122 & 0.407 & 0.501 \\
FLUX & 0.277 & 0.436 & 0.562 & \textcolor{brown!70!black}{0.844} & 0.848 & \textcolor{brown!70!black}{0.844} & \textcolor{gray!70!black}{0.550} & 12.282 & 0.373 & 0.586 \\
HQEdit & 0.316 & 0.578 & 0.711 & 0.774 & 0.668 & 0.670 & 0.220 & 8.952 & 0.291 & 0.420 \\
Step1X & \textcolor{orange!90!black}{0.247} & \textcolor{orange!90!black}{0.278} & \textcolor{orange!90!black}{0.411} & \textcolor{gray!70!black}{0.857} & \textcolor{orange!90!black}{0.878} & \textcolor{orange!90!black}{0.872} & \textcolor{brown!70!black}{0.510} & \textcolor{orange!90!black}{14.818} & \textcolor{orange!90!black}{0.515} & \textcolor{gray!70!black}{0.646} \\
\bottomrule
\end{tabular}}
\end{table*}

\section{Examples of Image Descriptions and Editing Instructions}
This section presents several image descriptions and editing instructions generated during the construction of Omni IIE Bench.

\begin{listing}[h]
\caption{Example description JSON in our benchmark.}
\begin{minted}[fontsize=\small, breaklines, frame=lines]{json}
{
    "replaceable_subject": {
        "name": "a hooded figure in a brown robe"
    },
    "modifiable_attributes": {
        "subject": [
            "robe is brown",
            "figure is hooded",
            "hands are clasped",
            "posture is upright"
        ],
        "background": [
            "background is dark and shadowy",
            "lighting is dramatic and focused on the figure"
        ]
    }
}
\end{minted}
\end{listing}

\begin{listing}[h]
\caption{Example description JSON in our benchmark.}
\begin{minted}[fontsize=\small, breaklines, frame=lines]{json}
{
    "replaceable_subject": {
        "name": "a brown bird with a long tail"
    },
    "modifiable_attributes": {
        "subject": [
            "feathers are brown",
            "tail is long and dark",
            "beak is curved and pointed",
            "legs are slender and gray"
        ],
        "background": [
            "floor is tiled and beige",
            "pillar is gray and cylindrical",
            "partial view of a black object (possibly a chair or table)"
        ]
    }
}
\end{minted}
\end{listing}

\begin{listing}[h]
\caption{Example description JSON in our benchmark.}
\begin{minted}[fontsize=\small, breaklines, frame=lines]{json}
{
    "replaceable_subject": {
        "name": "a modern bathroom interior"
    },
    "modifiable_attributes": {
        "subject": [
            "vanity cabinet is dark wood",
            "sink is white ceramic",
            "toilet is white ceramic",
            "shower curtain is beige with a floral pattern",
            "tub is white ceramic",
            "mirror is framed in gold"
        ],
        "background": [
            "walls are painted beige",
            "floor is light wood",
            "lighting is warm and ambient"
        ]
    }
}
\end{minted}
\end{listing}

\begin{listing}[h]
\caption{Example single-turn modification JSON in our benchmark.}
\begin{minted}[fontsize=\small, breaklines, frame=lines]{json}
{
    "modifications": [
        {
            "level": "low",
            "modification_text": "Change the color of the robe from brown to black."
        },
        {
            "level": "high",
            "modification_text": "Replace the hooded figure in a brown robe with a cloaked wizard wearing a pointy hat."
        }
    ]
}
\end{minted}
\end{listing}

\begin{listing}[h]
\caption{Example single-turn modification JSON in our benchmark.}
\begin{minted}[fontsize=\small, breaklines, frame=lines]{json}
{
    "modifications": [
        {
            "level": "low",
            "modification_text": "Change the feathers of the brown bird to be white."
        },
        {
            "level": "high",
            "modification_text": "Replace the brown bird with a gray rabbit."
        }
    ]
}
\end{minted}
\end{listing}

\begin{listing}[h]
\caption{Example multi-turn modification JSON in our benchmark.}
\begin{minted}[fontsize=\small, breaklines, frame=lines]{json}
{
    "modifications": [
        {
            "level": "1",
            "modification_text": "Change the color of the vanity cabinet to a lighter oak tone."
        },
        {
            "level": "2",
            "modification_text": "Replace the floral pattern of the shower curtain with a geometric pattern."
        },
        {
            "level": "3",
            "modification_text": "Change the lighting to a cooler, brighter tone."
        },
        {
            "level": "4",
            "modification_text": "Replace the gold frame of the mirror with a silver frame."
        },
        {
            "level": "5",
            "modification_text": "Change the color of the walls to a light gray."
        },
        {
            "level": "6",
            "modification_text": "Replace the white ceramic sink with a black granite sink."
        },
        {
            "level": "7",
            "modification_text": "Change the floor to a darker wood tone."
        },
        {
            "level": "8",
            "modification_text": "Replace the white ceramic toilet with a black ceramic toilet."
        },
        {
            "level": "9",
            "modification_text": "Change the white ceramic tub to a black granite tub."
        },
        {
            "level": "10",
            "modification_text": "Replace the light gray walls with a dark gray color."
        },
        {
            "level": "11",
            "modification_text": "Replace the vanity cabinet with a marble countertop and a black frame."
        },
        {
            "level": "12",
            "modification_text": "Replace the shower curtain with a clear glass shower door."
        }
    ]
}
\end{minted}
\end{listing}


\end{document}